# How Biomimetic Morphing Dorsal Fin Affects the Swimming Performance of a Free-swimming Tuna Robot

Hongbin Huang[1,2,3,4,†], Zhonglu Lin[1,2,†], Wei Zheng[1,2,3,4], Jinhu Zhang[1,2], Wei Zhou[3], Yu Zhang[1,2,*]

1 Key Laboratory of Underwater Acoustic Communication and Marine Information Technology of the Ministry of Education; Xiamen University.

2 State Key Laboratory of Marine Environmental Science, College of Ocean and Earth Sciences; Xiamen University.

3 Pen-Tung Sah Institute of Micro-Nano Science and Technology; Xiamen University.

4 College of Ocean and Earth Sciences and Discipline of Intelligent Instrument and Equipment; Xiamen University

† These authors contributed equally to this work
* Corresponding author. Email: yuzhang@xmu.edu.cn

**Abstract**
It is well known that tuna fish in the ocean can dynamically morph their median fins to achieve optimal hydrodynamic performance, e.g. linear acceleration and maneuverability. In this study, based on the previous studies about the median fin's hydrodynamic effects focusing on tethered conditions, we continue to explore the hydrodynamic function of tuna morphing dorsal fin in free swimming conditions for better approaching real-life situations. Here, we developed a tuna-inspired robotic fish platform that can swim independently in three dimensions, equipped with a biomimetic morphing dorsal fin magnetically attached to the robotic fish. Based on the free-swimming robotic fish platform, we investigated how the erected dorsal fin affects the speed, cost of transport (COT), and robotic fish's yaw angle at different frequencies and amplitudes. The erected dorsal fin plays a positive role in improving the yaw stability of robotic fish. However, it shows little influence on the speed and COT in our test. This remains to be further investigated in the future.

## 1 Introduction

In aquatic environments, fish are renowned for their exceptional locomotion performance. Over hundreds of millions of years, fish have evolved multi-fin systems that significantly contribute to their hydrodynamic performance[1]. Whether it is Body and Cauda Fin (BCF) or Median and Paired Fin (MPF) swimming mode fish, their median fins play a crucial role in propulsion, stability, and maneuverability of their locomotion[2]. For fish following the BCF swimming style, the median fin, particularly the dorsal fin, significantly improves the yaw stability[3], [4], swimming economy[5], motion acceleration[6], [7], and maneuverability[8].

Many researchers have been dedicated to investigating the mechanisms by which dorsal fins enhance the performance of robotic fish at tethered conditions. Zhong et al.'s [5] experimental research to examine the impact of dorsal fin sharpness on the swimming speed and efficiency of robotic fish. Han Pan et al. [9] discovered that the hydrodynamic interaction among fish's median fins yields a remarkable enhancement in propulsion efficiency, concurrently mitigating hydrodynamic resistance acting on the fish's body. Wen et al. [7] demonstrated that erecting the soft dorsal/anal fins increased linear acceleration. They subsequently introduced an "instantaneous self-propelled method" to eliminate the influence of guiding tracks and used it to characterize acceleration performance[6]. Li et al. [4] investigated the impact of various folding states of the median fin on the motion performance of tuna. These pioneering studies systematically elucidated the role of dorsal fins in fish locomotion through tethered condition experiments or numerical simulations.

In this study, we designed and fabricated a tuna-inspired robotic fish that can swim independently in three dimensions, equipped with a buoyancy control module, inertial measurement unit (IMU), and power sensor. Inspired by real tuna dorsal fins, we designed and built a bionic morphing dorsal fin magnetically attached to the body. Finally, we analyzed the effects of the erecting dorsal fin on speed, cost of transport (COT), and head yaw angle during swimming in a 3 × 2m pool. Compared with the previous studies, this work is the first free-swimming robotic fish equipped with a morphing dorsal fin, allowing a more precise and comprehensive analysis regarding the effects of morphing dorsal fin on fish swimming. The robotic fish developed in this study holds significant potential for engineering applications in the future. Leveraging the presence of morphing dorsal fin makes the robotic fish swim with increased agility, enhancing its adaptability to diverse environments.

## 2 Materials and method

### 2.1 System architecture

In the wild, the bluefin tuna can dynamically erect or fold its median fin according to its needs; this inspires the current robotic fish's morphing fin design, as seen in Figure 1A. Based on the appearance of the real tuna, the geometry is simplified by the curve of NACA0018—the specific dimensional details depicted in Figure 1B. The complete structure of the robotic fish includes the housing, microcontroller unit (MCU), computer, buoyancy control module, battery, morphing dorsal fin system, waterproof servo motor, and various sensors. The housing is manufactured with Polylactic Acid (PLA) through 3D printing. To prevent the softening and foaming caused by excessive soaking of PLA in water, it must be brushed with epoxy resin. The buoyancy control module uses a DC motor to drive the syringe piston to move and controls the weight of the whole machine slightly to adjust the buoyancy state. A lithium polymer battery powers all components. The robotic fish adopts a single-joint design and is driven by a waterproof servo motor. The sensors include an IMU, a depth sensor, and a direct current (DC) power sensor. The IMU is attached to the center of mass of the robotic fish and provides data on the acceleration and yaw angle of the robotic fish. The depth sensor transmits the current depth information of the robotic fish to the buoyancy control module. The DC power sensor is the INA219, which can measure the current and voltage consumed by the servo in real time while the fish robot swims and obtain the real-time power value through the integrated multiplier. Table 1 presents the primary technical parameters of the robotic fish.

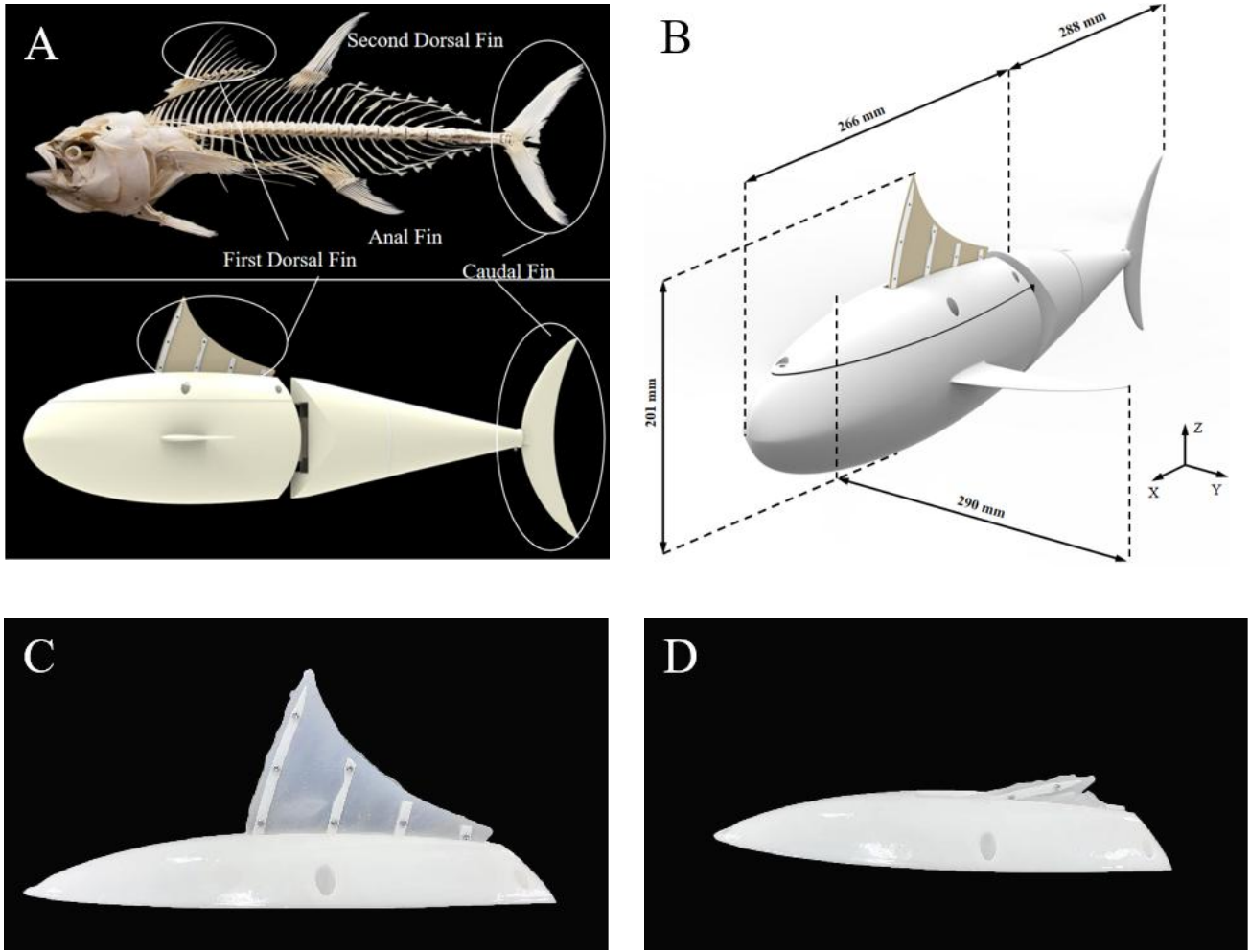

Figure 1. The bionic prototype and morphology of the bluefin tuna mid-fin system. (A) The skeletal model of bluefin tuna and the inspired overall design of the present paper (B) Detailed design of the current robotic tuna with dimensions. (C) Fully-erected Bionic morphing dorsal fins (D) Fully-folded Bionic morphing dorsal fins

**Table 1** The design specifications of the robotic fish

| Items | Specifications |
|---|---|
| Dimensions(mm) | 544(L)×290(W)×201(H) (Erect Dorsal Fin) |
|  | 544(L)×290(W)×128(H) (Fold Dorsal Fin) |
| Length of the head (mm) | 266 |
| Length of the tail (mm) | 288 |
| Mass (kg) | 2.305 |
| Maximum speed (mm/s) | 225(A=20° f=2.5Hz) |
| Microcontroller unit (MCU) | Raspberry Pi 4B |
| Servomotors | DG-3160 MG |
| Depth sensor | MSP20 |

## 2.2 Morphing dorsal fin design and manufacturing

The configuration and dimension of the morphing dorsal fin are derived from the actual tuna. Subsequently, the two-dimensional model is obtained by preserving essential features. The casting mold for the dorsal fin is obtained by a light-curing 3D printing process. Ecoflex00-30 is selected as the material for the dorsal fin casting, characterized by a tensile strength of 200 psi and a curing duration of 4 hours. The pictures of the fish's dorsal fin in fully erected and folded states are illustrated in Figures 1C and 1D, respectively.

To achieve the erected and folded states of the morphing dorsal fin system, it is necessary to integrate four fin rays within the dorsal fin mold. A four-bar linkage mechanism has been devised to facilitate the dorsal fin's erecting and folding motion, emulating an actual tuna. Magnets have been employed for affixing the base of each fin ray to the robotic fish's body. Influenced by the magnets, the base of the first fin ray can rotate under the influence of the internal magnetic field. Consequently, it controls the erecting and folding the entire dorsal fin through a four-bar linkage mechanism. This magnetic coupling addresses a fundamental challenge in self-propelled robotic fish design: achieving waterproofing.

## 2.3 Buoyancy control module

The swim bladder is a gas-filled sac-like organ in most teleost fishes. Its physiological role involves regulating the fish's buoyancy in water by altering its volume, thereby controlling displacement and buoyant forces to facilitate buoyancy adjustments for diving and surfacing [10]. The sketch of this principle is illustrated in Figure 2B.

Inspired by the swim bladder of fish, we designed and manufactured a buoyancy control module to enable the robotic fish to float up or dive down. The composition of the buoyancy control module is depicted in Figure 2A, which includes a syringe, piston, direct current (DC) motor, and gearbox. Equipped with a gearbox, the DC motor can reduce its turning speed, thereby increasing torque and effectively improving the reliability of water collection and drainage. To make the robotic fish close to the neutral buoyancy and sustain a horizontal attitude, the center of gravity is pre-adjusted by ballast weights to align with the center of buoyancy within the same vertical plane.

We employed a closed-loop PID control algorithm to enable the robotic fish to maintain a constant depth in the water. By controlling the symmetrical and asymmetrical swinging of the servo motor, it is possible to achieve both straight-line and turning movements for the robotic fish in a two-dimensional plane. We accomplish the upward and downward movements of the robotic fish through open-loop control of changes in the target depth, as depicted in Figures 2C and 2D. Furthermore, the robotic fish can effortlessly achieve free movement in three-dimensional space by integrating MCU control with servo motors for both symmetrical and asymmetrical oscillations.

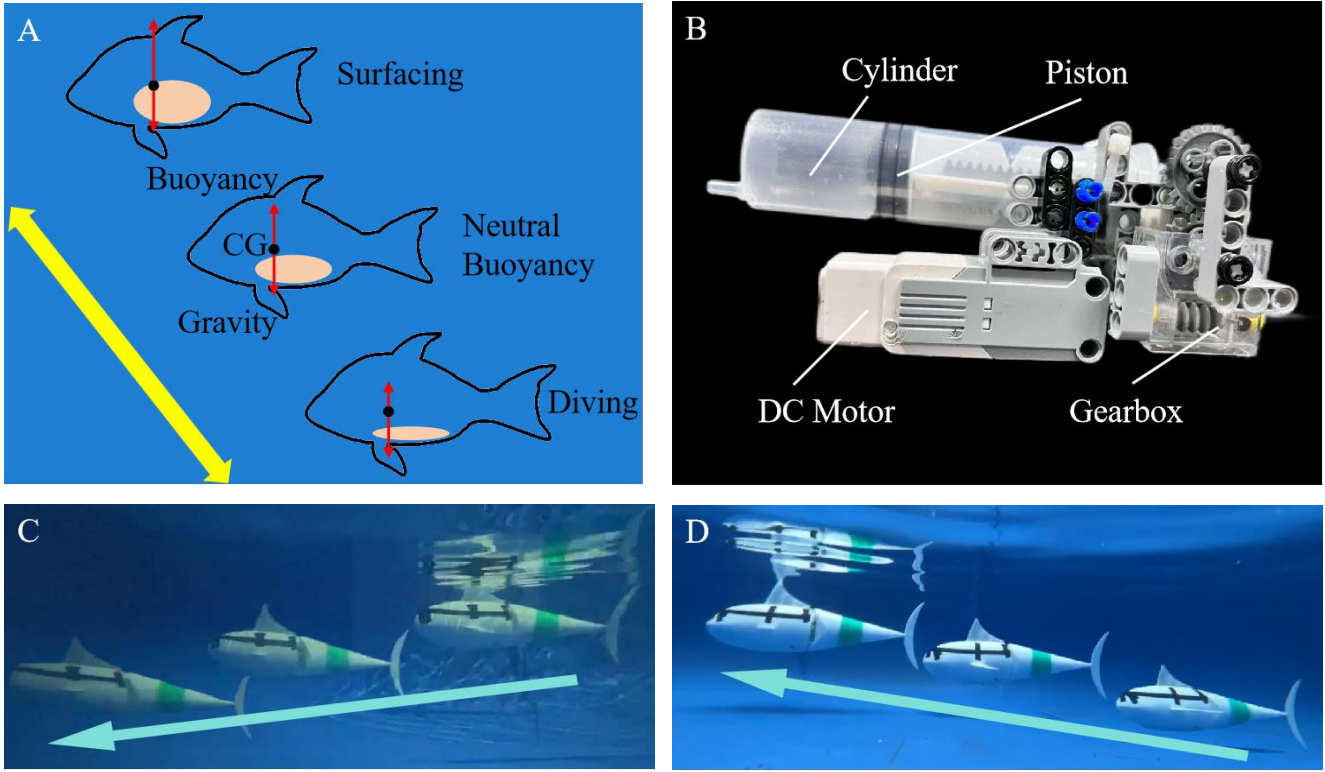

Figure 2. Three-dimensional movement of the robotic fish. (A) The buoyancy control module of the robotic fish. (B) The sketch of the surfacing and diving movement of the robotic fish. (C) Surfacing movement of the robotic fish. (D) Diving movement of the robotic fish.

## 3 Results and discussion

### 3.1 Speed and efficiency of the robotic fish

For robotic fish, speed and efficiency are critical measurements of their swimming ability, substantially influencing task completion duration and endurance. This section is dedicated to examining how the speed and efficiency of robotic fish vary under different caudal fin oscillation frequencies. The efficiency of a robotic fish swimming we generally define in terms of COT, calculated as:

$$\text{COT} = \frac{\overline{P_{in}}}{mg\overline{U}}$$

where $\overline{P_{in}}$ is the cycle-averaged input power of the robotic fish, measured by DC (direct current) power sensor INA219, m is the mass of the robotic fish, the acceleration of gravity g is equal to 9.81m/s², and $\overline{U}$ is the cycle-averaged speed of robotic fish.

The robotic fish accomplishes its linear underwater movement through a servo motor, which imparts a symmetrical oscillatory motion to the caudal fin. The experiments occurred in a 3×2m pool where the robotic fish swam straightforwardly from one pool end to the other. Upon covering a predetermined distance, the entire swimming process was captured via camera recording. The mean swimming velocity, denoted as $\overline{U}$, can be determined by dividing the pool's predefined length by the time taken to traverse it. The experiment involved the assessment of speed and power values at ten different frequencies (ranging from 0.8Hz to 2.33Hz), with five measurements taken for each frequency group and subsequent averaging of the recorded data, as seen in Figure 3. Subsequently, we establish the relationship between frequency and either linear velocities or COT. Our results reveal that the locomotion speed and power consumption of robotic fish, irrespective of the presence or absence of dorsal fins, varies in tandem with the oscillation frequency of the caudal fin. The experimental findings indicated an approximately quadratic correlation between the oscillation frequency of the caudal fin and the linear velocity of movement. At the highest frequency, the minimum COT values attained were 1.42 for the configuration with folded dorsal fin and 1.32 for the setup with erected dorsal fin.

Upon comparing the speed and COT profiles of the robotic fish swimming with and without the dorsal fin, it becomes evident that the dorsal fin exerts minimal influence on the speed and efficiency of its motions, particularly at lower speeds. This observation could be attributed to the constraints imposed by the experimental setting. It may necessitate further validation by testing the robotic fish in a significantly larger aquatic enclosure with a high speed.

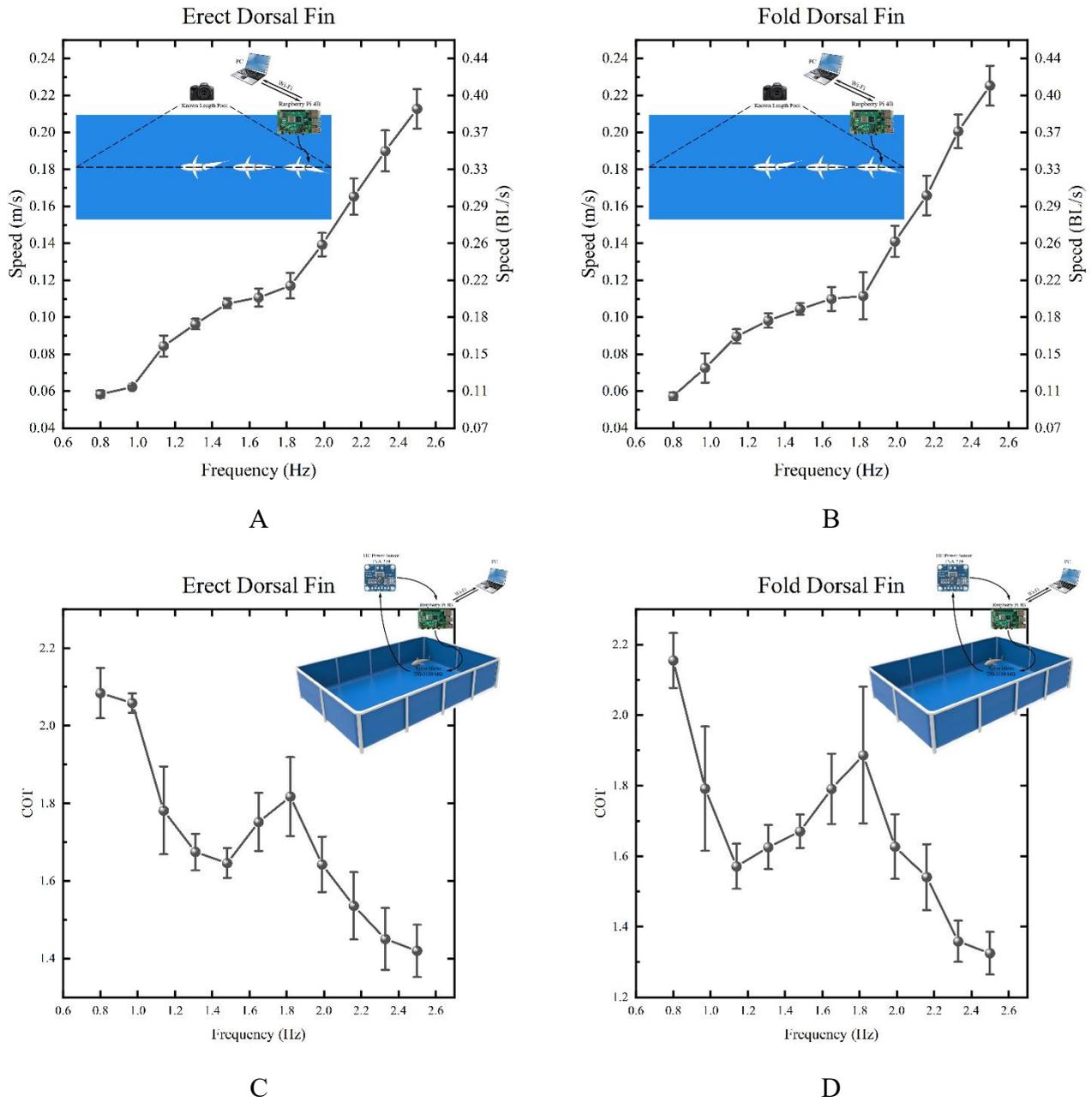

Figure 3. Study of robotic fish's linear motion and energy performance. Variation of average linear motion speed under different motion frequencies (A) when erecting the dorsal fin. (B) when folding the dorsal fin. Variation of COT under different motion frequencies when the dorsal fin is (C) erected or (D) folded.

**3.2 Yaw stability of the robotic fish**
Yaw stability has important implications for fish-like movement. The fluid-induced drag force of robotic fish can be calculated as follows:

$$F_d = -\frac{1}{2}\rho U^2 C_d A$$

where $F_d$ is the resistance of robotic fish, $\rho$ is the water density, $U$ is the speed of robotic fish compared with water, $C_d$ is the resistance coefficient, and A is the waterfront area. As the yaw angle of the fish's head becomes more pronounced, there is a corresponding increase in the resistance encountered by the fish, which consequently impacts its overall movement performance. In the natural environment, fish mitigate this effect by employing muscular control to reduce the yaw angle of their heads. Nonetheless, the absence of suitable control mechanisms in robotic fish constitutes a significant factor contributing to their notable lag behind their biological counterparts regarding locomotion capabilities.

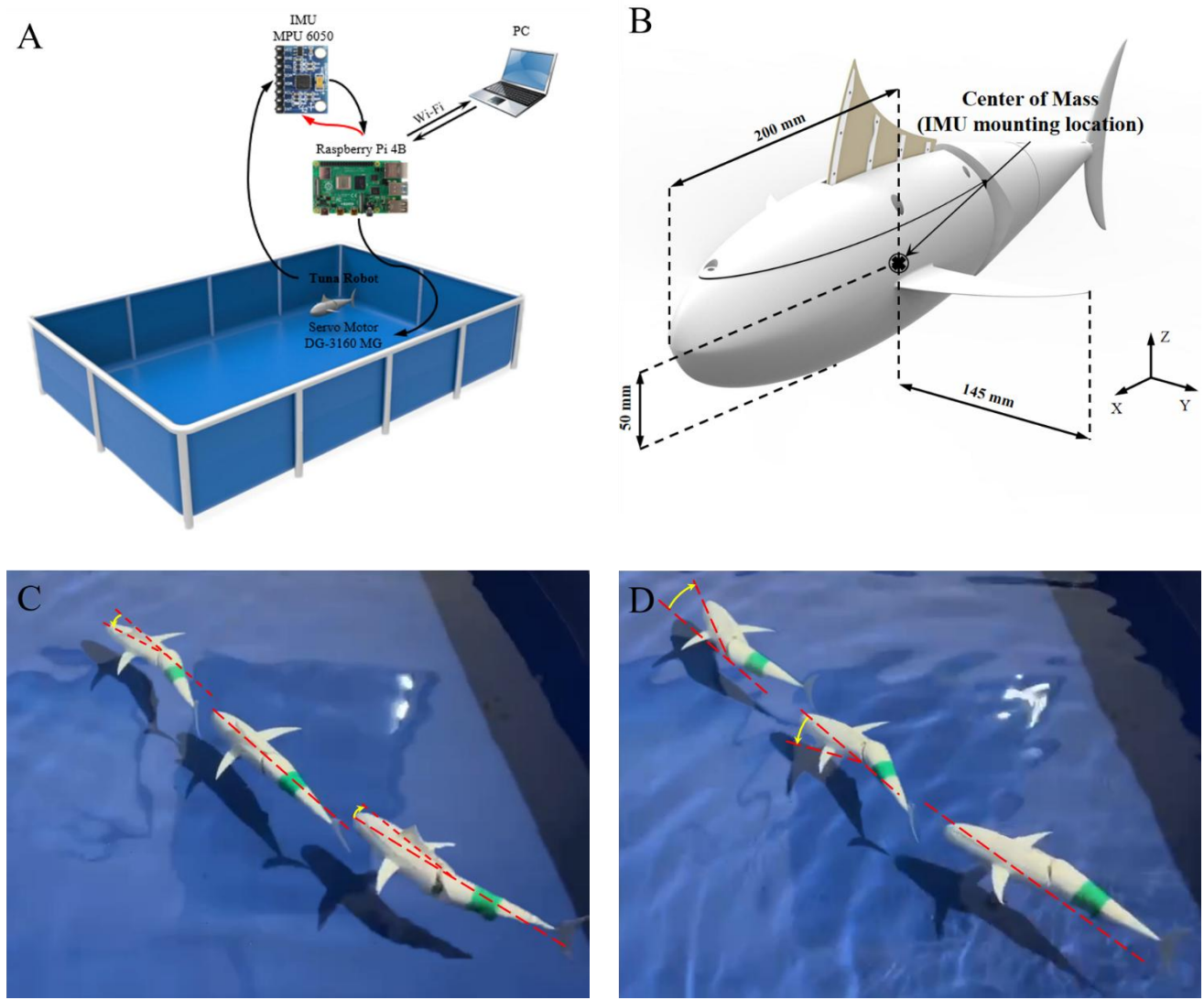

Figure 4. Robotic fish yawing motion experiment. (A) The setup of the robotic fish yawing motion experiment. (B) Experiment platform of robotic fish swimming. (C) Screenshot of robotic fish swimming with erected dorsal fin. (D) Screenshot of robotic fish swimming with folded dorsal fin.

The experimental setup is depicted in Figure 4A, comprising a 3×2m pool, the robotic fish, and a computer. The MCU of the robotic fish is Raspberry Pi 4B, which is connected to the computer through WIFI. The MCU is responsible for controlling the movement of the robot fish and recording the experimental data. The MCU controls the robotic fish's movement and records the IMU's experimental data as an internal storage unit. As shown in Figure 4B, to reflect the motion of the robotic fish as authentically as possible, the IMU is positioned close to the center of mass.

Throughout the experiment, various amplitudes and frequencies of the robotic fish's caudal fin were configured, and the IMU was utilized to capture the yaw angle of the robotic fish's head for each specific configuration.

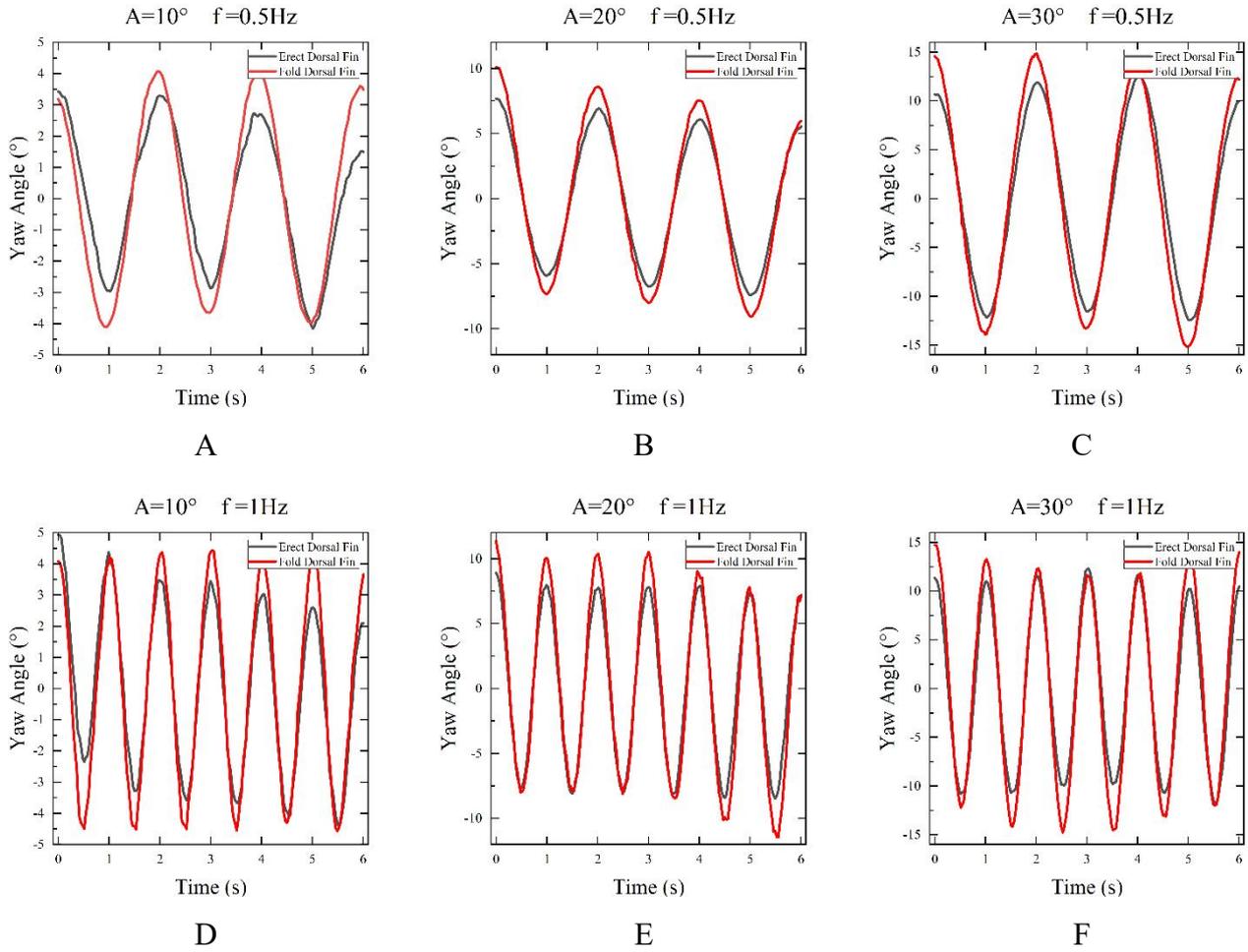

Figure 5. Yaw angle of the robotic fish at different oscillation amplitudes and frequencies of the caudal fin. (A) oscillation amplitude of 10° with a frequency of 1 Hz. (B) oscillation amplitude of 20° with a frequency of 1 Hz. (C) oscillation amplitude of 30° with a frequency of 1 Hz. (D) oscillation amplitude of 10° with a frequency of 0.5 Hz. (E) oscillation amplitude of 20° with a frequency of 0.5 Hz. (F) oscillation amplitude of 30° with a frequency of 0.5Hz.

In this experiment, we set the caudal fin oscillation frequency to 0.5 Hz and 1 Hz, and the oscillation amplitude to 10°, 20°, and 30°, with the dorsal fin erected and the dorsal fin folded, respectively. As the amplitude of the caudal fin swing increases, the yaw angle of the robotic fish becomes larger. As shown in Figure 5, the erected dorsal fin can suppress the head yaw of the robotic fish. Table 2 summarizes the yaw angle amplitude of the robotic fish's head for all cases. The erected dorsal fin reduced the peak-to-peak amplitude of the yaw angle from 18.47° to 14.01°, with a maximum reduction of 24.20%, when the caudal fin oscillated at a frequency of 1 Hz and an amplitude of 20°.

Table 2 Comparison of peak-to-peak amplitudes of yaw angle

| Amplitudes-Frequency | With Dorsal Fin(°) | Without Dorsal Fin (°) | Improvement of Yaw Stability |
|---|---|---|---|
| 10°-0.5Hz | 6.07 | 7.65 | 20.67% |
| 10°-1.0Hz | 6.99 | 8.64 | 19.12% |
| 20°-0.5Hz | 13.24 | 16.16 | 18.14% |
| 20°-1.0Hz | 14.01 | 18.47 | 24.20% |
| 30°-0.5Hz | 23.32 | 27.93 | 16.51% |
| 30°-1.0Hz | 21.85 | 26.47 | 17.47% |

**4 Conclusions**

In this study, a free-swimming robotic fish was designed and manufactured, inheriting a biomimetic morphing dorsal fin, a buoyancy control module, and wireless communication electronics. This robotic fish can free three-dimensional motion within a laboratory aquatic environment. We systematically measured the velocity, COT, and head yaw angle through experimental procedures in both the dorsal fin erected and folded states. The results indicate that the erected dorsal fin has a significantly positive effect on improving the robotic fish's yaw stability. However, despite suppression of head yaw, the morphing dorsal fin exhibits little influence on the speed and COT in our tested range of parameters. This remains to be further investigated in the future.

2020, doi: 10.1115/1.4045901.